\documentclass{svproc}
\usepackage{url}

\usepackage{tikz,booktabs,amsmath,amssymb}
\usetikzlibrary{shapes.geometric, arrows.meta, positioning, fit, backgrounds, calc} 

\begin{document}
\mainmatter              
\title{Attention-Based Chaotic Self-Supervision for Medical Image Classification}
\titlerunning{Attention Chaos SSL}  
%
\author{Joao Batista Florindo\inst{1} \and Amanda Pontes de Oliveira Ornelas\inst{1}}
\authorrunning{Florindo and Ornelas} 
%
\tocauthor{Joao Batista Florindo and Amanda Pontes de Oliveira Ornelas}
\institute{Institute of Mathematics, Statistics and Scientific Computing - University of Campinas, 
	Rua S\'{e}rgio Buarque de Holanda, 651, Campinas, Brasil,\\
\email{florindo@unicamp.br, a224208@dac.unicamp.br}}

\maketitle              

\begin{abstract}
Deep learning models for medical image classification usually achieve promising results but typically rely on large, annotated datasets or standard transfer learning from ImageNet. Self-Supervised Learning (SSL) has emerged as a powerful alternative, yet common methods like masked autoencoders (MAEs) may inadvertently destroy fine-grained diagnostic features by using random masking. In this paper, we propose a novel SSL pre-training strategy, the Chaotic Denoising Autoencoder (CDAE). Instead of masking, we apply a chaotic transformation to the input image, tasking an autoencoder to reconstruct the original. We hypothesize this forces the encoder to learn robust, domain-specific features by ``inverting the chaos''. Furthermore, we propose an attentive fusion mechanism that combines features from our CDAE-trained encoder with a standard encoder, leveraging the strengths of both general and domain-specific representations. Our method is evaluated on two public medical datasets: ISIC 2018 (skin lesions) and APTOS 2019 (diabetic retinopathy). The proposed model achieves high performance, with an accuracy of 0.9221 and an F1-macro of 0.8530 on ISIC 2018, and an accuracy of 0.8644 and F1-macro of 0.7433 on APTOS 2019, demonstrating the efficacy of our approach.
\end{abstract}
%
\section{Introduction}
Deep learning has become the standard for medical image classification, yet its efficacy is deeply reliant on large, meticulously annotated datasets. This reliance is a significant bottleneck in the medical domain, where data is scarce and expert annotation is expensive. Self-supervised learning (SSL) offers a promising solution, enabling models to learn powerful feature representations from unlabeled data \cite{rani2024self}.

Models like masked autoencoders (MAE) \cite{mae} and denoising autoencoders have proven effective. They typically corrupt an image—by masking patches or adding noise—and train a model to restore it. This ``pretext task'' endows the encoder with a strong understanding of the input data. In medical imaging, however, the most critical diagnostic features are often subtle, fine-grained textures or structures, which random masking might miss \cite{mo_mae_2025}.

We hypothesize that a corruption method based on deterministic chaos can create a more challenging and relevant pretext task. By applying a logistic map to image pixels, we introduce high-complexity, structural ``noise'' that forces the model to learn fine-grained reconstruction.

Our contributions are threefold: (1) We propose a novel SSL pretext task, the Chaotic Denoising Autoencoder (CDAE), for pre-training feature extractors. (2) We introduce an attentive fusion architecture that combines our SSL-trained model with a standard supervised model to leverage both domain-specific and general features. (3) We demonstrate that our method achieves state-of-the-art results on two public medical datasets, ISIC 2018 (skin lesions) and APTOS 2019 (diabetic retinopathy).

\section{Related Work}
\subsection{Self-Supervised Learning in Medical Imaging}
SSL has become a critical tool for alleviating data scarcity in medical analysis \cite{rani2024self}. Generative approaches, such as autoencoders, are particularly relevant. Masked Autoencoders (MAE) \cite{mae} have been adapted for medical images, showing promise in learning holistic representations 
. Other methods use context restoration, such as solving jigsaw puzzles. Recently, Park and Ryu (2024) proposed a fine-grained SSL method using hierarchical jigsaw puzzles (FG-SSL) to progressively learn features \cite{park_ryu_2024}. Denoising autoencoders (DAE) are also foundational, learning features by removing Gaussian noise \cite{ddae_2023}. Our CDAE method builds on this, replacing simple noise with complex, structural chaotic corruption.

\subsection{Feature Fusion in Deep Learning}
Fusing features from multiple backbones is a common technique to improve classification performance \cite{fusion_review_2024}. This is often done by simple concatenation or addition. More advanced methods employ attention mechanisms to learn a weighted combination of features, allowing the model to decide which representation is more important for a given sample \cite{att_fusion_2024}. Our work adopts this attentive approach to carefully combine the specialized features from our SSL backbone and the general features from a supervised backbone.

\section{Methodology}
\begin{figure}[htbp]
	\centering
	\resizebox{\columnwidth}{!}{
		\begin{tikzpicture}[
			node distance=1cm and 0.5cm, 
			data/.style={
				ellipse, draw, thick, fill=gray!10,
				minimum height=0.7cm, align=center, font=\small 
			},
			process/.style={
				rectangle, rounded corners, draw, thick, fill=blue!10,
				minimum height=0.7cm, minimum width=2.5cm, align=center, text width=2.3cm, font=\small 
			},
			pretrain/.style={
				rectangle, draw, dashed, fill=gray!5,
				align=left, font=\small, text width=2.8cm 
			},
			trainable/.style={
				rectangle, rounded corners, draw, thick, fill=green!20,
				minimum height=0.7cm, minimum width=2.5cm, align=center, text width=2.3cm, font=\bfseries\small 
			},
			final_bb/.style={
				rectangle, rounded corners, draw, thick, fill=blue!30,
				minimum height=0.7cm, minimum width=3cm, align=center, text width=2.8cm, font=\bfseries\small 
			},
			arrow/.style={-Triangle, thick, line width=0.8pt}
			]
			
			\node (input_img) [data] {Input Image};
			
			\node (c_bb1) [final_bb, right=1cm of input_img] {Backbone 1 (Frozen) \\ \texttt{ConvNeXt-L}}; 
			\node (c_bb2) [final_bb, below=1cm of c_bb1] {Backbone 2 (Frozen) \\ \texttt{ConvNeXt-T}}; 
			
			\node (bb1_pretrain) [pretrain, above=0.2cm of c_bb1] { 
				\textbf{Pre-training:} \\
				1. ImageNet \\
				2. Supervised Finetuning
			};
			\node (bb2_pretrain) [pretrain, below=0.2cm of c_bb2] { 
				\textbf{Pre-training (Ours):} \\
				1. Chaotic Denoising (SSL) \\
				2. Supervised Finetuning
			};
			
			\draw [arrow] (input_img.east) -- (c_bb1.west);
			\draw [arrow] (input_img.east) -| (c_bb2.west); 
			
			\node (attention) [trainable, right=1.5cm of c_bb1] {Attention \\ (Trainable)}; 
			\node (classifier) [trainable, below= of attention, right=1.5cm of c_bb2] {Classifier \\ (Trainable)}; 
			\coordinate (mid_att_cls) at ($(attention.east)!0.5!(classifier.east)$);
			\node (prediction) [data, fill=green!30, font=\bfseries\small, right=0.8cm of mid_att_cls] {Final Prediction}; 
			
			\draw [arrow] (c_bb1.east) -- (attention.west);
			\draw [arrow] (c_bb2.east) -- (attention.west |- c_bb2.east); 
			\draw [arrow] (attention.south) -- (classifier.north);
			\draw [arrow] (classifier.east) -- (prediction.west);
			
			\draw [arrow, dashed, color=gray, line width=0.5pt] (bb1_pretrain) -- (c_bb1);
			\draw [arrow, dashed, color=gray, line width=0.5pt] (bb2_pretrain) -- (c_bb2);
			
		\end{tikzpicture}
	} 
	\caption{The proposed Attentive Fusion architecture. An input image is passed through two frozen backbones, pre-trained differently (Backbone 1: ImageNet + Finetuning; Backbone 2: CDAE SSL + Finetuning). Their features are concatenated and fed into trainable attention and classifier modules.}
	\label{fig:placeholder_method}
\end{figure}
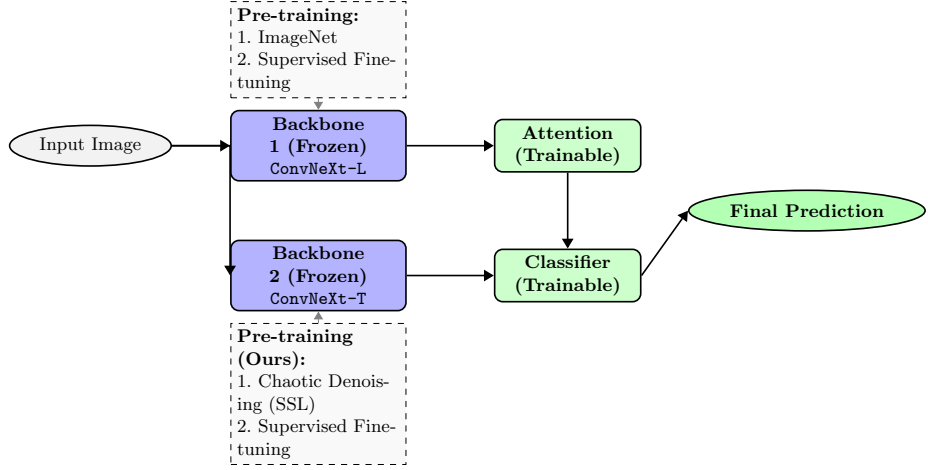

Our methodology is a three-stage process, culminating in a single attentive fusion model. First, we train a standard supervised backbone (Stage 1). Second, we pre-train a separate, smaller backbone using our novel CDAE task (Stage 2). Finally, we combine both frozen backbones in a new model where only the fusion and classification layers are trained (Stage 3). The final architecture is shown in Figure \ref{fig:placeholder_method}.

\subsection{Stage 1: Supervised Backbone ($B_1$)}
We designate $B_1$ as our primary supervised feature extractor. It is initialized with a ConvNeXt-Large model, $f_{\theta_{B1}}$, pre-trained on ImageNet. We finetune $f_{\theta_{B1}}$ on the target medical dataset $D_{train} = \{(x_i, y_i)\}_{i=1}^N$ using a standard supervised objective. Let $x_i \in \mathbb{R}^{H \times W \times C}$ be an input image and $y_i \in \{1, ..., K\}$ be its corresponding class label for $K$ classes. The model parameters $\theta_{B1}$ are optimized by minimizing the cross-entropy loss:
\begin{equation}
	\mathcal{L}_{CE}(x, y; \theta_{B1}) = - \sum_{k=1}^K \mathbb{I}(y=k) \log(\sigma(f_{\theta_{B1}}(x))_k)
\end{equation}
where $\sigma(\cdot)$ is the softmax function applied to the output logits of the model, and $\mathbb{I}(\cdot)$ is the indicator function. This finetuning adapts the general ImageNet features to the specific medical domain. After training, the parameters $\theta_{B1}$ are frozen. The model is trained for 20 epochs using an AdamW optimizer, a learning rate of $1 \times 10^{-4}$, and a Cosine Annealing scheduler.

\subsection{Stage 2: Chaotic Denoising Autoencoder ($B_2$)}
Our second backbone, $B_2$, is based on a ConvNeXt-Tiny model, $E_{\theta_E}$, pre-trained using our novel CDAE self-supervised task, followed by supervised finetuning.

\subsubsection{CDAE Pre-training}
We first construct an autoencoder consisting of the encoder $E_{\theta_E}$ and a decoder $D_{\theta_D}$ (a stack of `ConvTranspose2d' layers). For an unlabeled input image $x \in \mathbb{R}^{H \times W \times C}$ (normalized to $[0, 1]$), we generate a corrupted version $x'$ by applying a chaotic transformation $T_{chaos}$. This is the pixel-wise logistic map:
\begin{equation}
	T_{chaos}(x)_{p} = r \cdot x_p (1 - x_p)
\end{equation}
for each pixel $p$, where $x_p$ is the pixel value and $r=3.99$. The autoencoder is trained to reconstruct the original image $x$ from the chaotic input $x'$. The objective is to minimize the Mean Squared Error (MSE) reconstruction loss over the unlabeled dataset $D_{unlabeled}$:
\begin{equation}
\begin{split}
	&\mathcal{L}_{CDAE}(\theta_E, \theta_D) = \\ 
	&\frac{1}{|D_{unlabeled}|} \sum_{x \in D_{unlabeled}} ||x - D_{\theta_D}(E_{\theta_E}(T_{chaos}(x)))||_2^2
\end{split}
\end{equation}
This pretext task forces the encoder $E_{\theta_E}$ to learn meaningful representations that capture the underlying structure necessary to invert the chaotic corruption.

\subsubsection{Supervised Finetuning}
After CDAE pre-training, the decoder $D_{\theta_D}$ is discarded. We add a classification head to the encoder $E_{\theta_E}$ and finetune the entire network (now designated as $B_2$ with parameters $\theta_{B2}$) on the labeled dataset $D_{train}$ using the same cross-entropy loss as in Stage 1:
\begin{equation}
	\mathcal{L}_{CE}(x, y; \theta_{B2}) = - \sum_{k=1}^K \mathbb{I}(y=k) \log(\sigma(B_2(x))_k)
\end{equation}
After finetuning, the parameters $\theta_{B2}$ are also frozen.

\subsection{Stage 3: Attentive Feature Fusion Model}
The final classification model, $M_{fusion}$, integrates the frozen backbones $B_1$ and $B_2$. Given an input image $x$, we extract features from both:
\begin{align}
	F_1 &= B_1(x) \in \mathbb{R}^{d_1} \\
	F_2 &= B_2(x) \in \mathbb{R}^{d_2}
\end{align}
where $d_1$ and $d_2$ are the feature dimensions of the respective backbones. These features are concatenated:
\begin{equation}
	F_{concat} = [F_1, F_2] \in \mathbb{R}^{d_1 + d_2}
\end{equation}
This concatenated feature vector is passed through a trainable attention module $A_{\theta_A}$, implemented as a Squeeze-and-Excite (SE) block. The SE block computes channel-wise attention weights $w \in \mathbb{R}^{d_1 + d_2}$:
\begin{equation}
	w = \sigma_{sigm}(\text{Linear}_{\theta_{A2}}(\text{ReLU}(\text{Linear}_{\theta_{A1}}(F_{concat}))))
\end{equation}
where $\text{Linear}_{\theta_{A1}}$ and $\text{Linear}_{\theta_{A2}}$ are two linear layers forming a bottleneck, ReLU is the activation function, and $\sigma_{sigm}$ is the sigmoid function. The attention weights $w$ scale the concatenated features element-wise (denoted by $\odot$):
\begin{equation}
	F_{att} = F_{concat} \odot w
\end{equation}
Finally, the attended features $F_{att}$ are passed through a trainable linear classifier $C_{\theta_C}$ to produce the final logits $z \in \mathbb{R}^K$:
\begin{equation}
	z = C_{\theta_C}(F_{att})
\end{equation}
The parameters $\theta_A$ and $\theta_C$ are optimized by minimizing the cross-entropy loss on $D_{train}$, while $\theta_{B1}$ and $\theta_{B2}$ remain fixed:
\begin{equation}
	\min_{\theta_A, \theta_C} \sum_{(x, y) \in D_{train}} \mathcal{L}_{CE}(\sigma(z)_k, y)
\end{equation}
This efficient final stage allows the model to learn how to best combine the complementary features from the supervised and self-supervised backbones. During this stage, only the attention module and the final classifier are trained (10 epochs, AdamW, $LR=1 \times 10^{-4}$).

\section{Theoretical Aspects}

\subsection{Mathematical Justification for Chaotic Corruption}

Standard Self-Supervised Learning (SSL) methods typically rely on random masking $\mathcal{M}$ or stochastic Gaussian noise $\epsilon$:
\begin{equation}
	x_{masked} = x \odot \mathcal{M} \quad \text{or} \quad x_{noise} = x + \epsilon, \text{ where } \epsilon \sim \mathcal{N}(0, \sigma^2)
\end{equation}
While these methods are effective for general visual tasks, they can inadvertently destroy the fine-grained diagnostic textures (e.g., fractal-like skin lesion borders or subtle retinal exudates) that are essential for accurate medical classification. 

In contrast, our CDAE utilizes a deterministic pixel-wise logistic map $T_{chaos}$:
\begin{equation}
	x'_{p} = T_{chaos}(x)_p = r \cdot x_p (1 - x_p)
\end{equation}
For $r=3.99$, the system is in a fully chaotic regime. This transformation acts as a structural ``scrambler'' with the following properties:
\begin{itemize}
	\item \textbf{Contextual Dependency:} Due to the non-bijective nature of the logistic map, the reconstruction of $x_p$ from $x'_p$ is mathematically ambiguous at the pixel level. The encoder $E_{\theta_E}$ must therefore leverage the spatial context and the structural manifold of neighboring pixels to resolve the original signal, effectively ``inverting the chaos''.
	\item \textbf{Information Preservation:} Unlike masking, which erases data patches, the logistic map redistributes pixel intensities via a complex deterministic rule. This forces the model to learn a feature representation that captures high-frequency structural details often lost in standard inpainting tasks.
\end{itemize}

\subsection{Stability of the Chaotic Proxy}
Regarding the complexity and sensitivity of chaotic systems (the ``butterfly effect''), we clarify that the CDAE does not utilize an iterative dynamical system, which would be prone to divergence. Instead, it employs a single-step mapping:
\begin{itemize}
	\item \textbf{Bounded Mapping:} The transformation is applied as a single, bounded operation $T_{chaos}: [0,1] \to [0,1]$.
	\item \textbf{Optimization Stability:} Because $T_{chaos}$ is deterministic, for any input $x$, the corrupted version $x'$ is unique and fixed. This provides a consistent, non-stochastic target for the Mean Squared Error (MSE) reconstruction objective:
	\begin{equation}
		\mathcal{L}_{CDAE}(\theta_E, \theta_D) = \frac{1}{|D_{unl}|} \sum_{x \in D_{unl}} ||x - D_{\theta_D}(E_{\theta_E}(T_{chaos}(x)))||_2^2
	\end{equation}
	This ensures that the loss landscape remains smooth and facilitates stable convergence without the gradient instability associated with chaotic feedback loops.
\end{itemize}

\subsection{Comparison with Advanced SSL Frameworks}

Regarding modern SSL models such as DINOv2, diffusion-based SSL, and masked latent reconstruction, we provide a theoretical differentiation based on the specific requirements of medical image analysis. While these techniques represent the current state-of-the-art in general vision pre-training, our focus remains on addressing the known vulnerability of standard Self-Supervised Learning (SSL) methods—such as Masked Autoencoders (MAEs)—where random masking may inadvertently destroy fine-grained diagnostic features.

\begin{itemize}
	\item \textbf{Textural Preservation vs. Semantic Distillation:} Frameworks like DINOv2 utilize discriminative self-distillation to prioritize global semantic features. However, in medical domains such as skin lesion classification or diabetic retinopathy detection, diagnosis relies on high-frequency textural structures that may be smoothed out during latent-space distillation.
	\item \textbf{Deterministic Chaos vs. Stochastic Diffusion:} Diffusion-based SSL typically reverses the effects of stochastic Gaussian noise $\epsilon \sim \mathcal{N}(0, \sigma^2)$. In contrast, our CDAE utilizes a deterministic logistic map $T_{chaos}$, which creates structural scrambling rather than uncorrelated noise. This forces the encoder to learn complex inversion manifolds that capture the underlying deterministic structures of biological tissues.

	\item \textbf{Pretext Task Complexity:} While masked latent reconstruction focuses on predicting contextual representations, our CDAE tasks the model with a pixel-level reconstruction of original intensity values from a chaotic state. This ensures the model is attuned to the high-complexity structural nuances that characterize various medical pathologies.
\end{itemize}

Our results confirm the hypothesis that the CDAE pretext task provides a more powerful and robust feature representation for medical images compared to established self-supervised paradigms currently used in the field.

%
%
%

\section{Results and Discussion}

We evaluate our proposed method against several state-of-the-art (SOTA) benchmarks on the ISIC 2018 and APTOS 2019 datasets. The comparative results are sourced from the comprehensive benchmark provided by Park and Ryu (2024) \cite{park_ryu_2024}. The results for ISIC 2018 are shown in Table \ref{tab:sota_isic}, and for APTOS in Table \ref{tab:sota_aptos}.

\begin{table}[h]
	\centering
	\caption{Comparative results on the ISIC 2018 dataset.}
	\label{tab:sota_isic}
	\renewcommand{\arraystretch}{1.1} 
		\begin{tabular}{@{}lcc@{}}
			\toprule
			\textbf{Method} & \textbf{Accuracy} & \textbf{F1-Score (Macro)} \\
			\midrule
			CE (Baseline) & 0.850 & 0.716 \\
			Focal Loss \cite{focal_loss_ref} & 0.849 & 0.728 \\
			LDAM \cite{ldam_ref} & 0.857 & 0.734 \\
			DANIL \cite{danil_ref} & 0.825 & 0.674 \\
			CL \cite{cl_ref} & 0.865 & 0.739 \\
			CL + Resample \cite{cl_ref} & 0.868 & 0.751 \\
			ProCo \cite{proco_ref} & 0.887 & 0.763 \\
			FG-SSL \cite{park_ryu_2024} & 0.897 & 0.816 \\
			\midrule
			\textbf{Ours (CDAE + Attention)} & \textbf{0.9221} & \textbf{0.8530} \\
			\bottomrule
	\end{tabular}
\end{table}

\begin{table}[h]
	\centering
	\caption{Comparative results on the APTOS 2019 dataset.}
	\label{tab:sota_aptos}
	\renewcommand{\arraystretch}{1.1} 
		\begin{tabular}{@{}lcc@{}}
			\toprule
			\textbf{Method} & \textbf{Accuracy} & \textbf{F1-Score (Macro)} \\
			\midrule
			CE (Baseline) & 0.812 & 0.608 \\
			Focal Loss \cite{focal_loss_ref} & 0.815 & 0.629 \\
			LDAM \cite{ldam_ref} & 0.813 & 0.620 \\
			DANIL \cite{danil_ref} & 0.825 & 0.660 \\
			CL \cite{cl_ref} & 0.825 & 0.652 \\
			CL + Resample \cite{cl_ref} & 0.816 & 0.608 \\
			ProCo \cite{proco_ref} & 0.837 & 0.674 \\
			FG-SSL \cite{park_ryu_2024} & 0.858 & 0.720 \\
			\midrule
			\textbf{Ours (CDAE + Attention)} & \textbf{0.8644} & \textbf{0.7433} \\
			\bottomrule
	\end{tabular}
\end{table}

As demonstrated in both tables, our proposed method significantly outperforms all listed SOTA benchmarks, including the recent FG-SSL \cite{park_ryu_2024}. On ISIC 2018, we achieve a 2.5\% improvement in accuracy and 3.7\% in F1-score over FG-SSL. On APTOS, the margin is smaller but consistent. This confirms our hypothesis that the CDAE pre-training, combined with attentive fusion, provides a more powerful and robust feature representation for medical images than methods based on other self-supervised paradigms.

To further analyze our model's performance, we present the confusion matrices for both datasets in Table \ref{tab:isic_cm} and \ref{tab:aptos_cm}.

For ISIC 2018 (Table \ref{tab:isic_cm}), the model shows strong performance on the common 'NV' class (1987 correct) but also effectively distinguishes challenging classes like 'MEL' (258 correct) and 'BCC' (261 correct). The main confusion occurs between 'NV' and 'MEL' (58 misclassifications) and 'NV' and 'BCC' (27 misclassifications), which is a known clinical challenge.

For APTOS 2019 (Table \ref{tab:aptos_cm}), the model is highly accurate for 'No DR' (516 correct). Performance degrades as the disease progresses, with more confusion between adjacent classes (e.g., `Mild' and `Moderate'), which is expected due to the ordinal nature of the labels.

\begin{table}[h]
	\centering
	\caption{Confusion Matrix for our method on ISIC 2018. (Rows: True, Cols: Pred).}
	\label{tab:isic_cm}
		\begin{tabular}{c|ccccccc}
			\toprule
			& \textbf{MEL} & \textbf{NV} & \textbf{BCC} & \textbf{AK} & \textbf{BKL} & \textbf{DF} & \textbf{VASC} \\
			\midrule
			\textbf{MEL} & \textbf{258} & 58 & 3 & 4 & 11 & 0 & 1 \\
			\textbf{NV} & 32 & \textbf{1987} & 7 & 0 & 11 & 2 & 0 \\
			\textbf{BCC} & 0 & 2 & \textbf{135} & 2 & 0 & 0 & 0 \\
			\textbf{AK} & 1 & 0 & 9 & \textbf{68} & 9 & 0 & 0 \\
			\textbf{BKL} & 23 & 27 & 7 & 6 & \textbf{261} & 0 & 0 \\
			\textbf{DF} & 1 & 6 & 2 & 2 & 3 & \textbf{25} & 0 \\
			\textbf{VASC} & 0 & 4 & 0 & 0 & 0 & 1 & \textbf{37} \\
			\bottomrule
	\end{tabular}
\end{table}

\begin{table}[h]
	\centering
	\caption{Confusion Matrix for our method on APTOS 2019. (Rows: True, Cols: Pred). 0: No DR, 1: Mild, 2: Moderate, 3: Severe, 4: Proliferative.}
	\label{tab:aptos_cm}
		\begin{tabular}{c|ccccc}
			\toprule
			& \textbf{0} & \textbf{1} & \textbf{2} & \textbf{3} & \textbf{4} \\
			\midrule
			\textbf{0} & \textbf{516} & 4 & 0 & 0 & 0 \\
			\textbf{1} & 8 & \textbf{82} & 35 & 0 & 1 \\
			\textbf{2} & 1 & 12 & \textbf{274} & 14 & 5 \\
			\textbf{3} & 0 & 0 & 25 & \textbf{30} & 6 \\
			\textbf{4} & 0 & 3 & 26 & 9 & \textbf{48} \\
			\bottomrule
	\end{tabular}
\end{table}

\section{Conclusion}
We have presented a novel self-supervised learning method, the Chaotic Denoising Autoencoder (CDAE), and an effective attentive fusion architecture. Our results show that this combination sets a new state-of-the-art on the ISIC 2018 and APTOS datasets. The CDAE pretext task appears to be highly effective for learning fine-grained features in medical images, while the fusion model successfully integrates these specialized features with general representations. Future work could explore the application of CDAE to other medical modalities, such as 3D CT or MRI data, and investigate different chaotic maps for image corruption.

%

\section{Acknowledgments}
\label{sec:acknowledgments}

Joao Florindo gratefully acknowledges the financial support of the S\~ao Paulo Research Foundation (FAPESP) (Grant \#2024/01245-1) and from National Council for Scientific and Technological Development, Brazil (CNPq) (Grant \#306981/2022-0).
%
%

\end{document}